\documentclass[11pt,twoside]{article}
\usepackage{geometry}
\usepackage{fancyhdr}
\fancypagestyle{firststyle}
{
   \fancyhf{}
   \fancyfoot{}
}

\usepackage{latexsym}
\usepackage{todonotes}
\usepackage[latin1]{inputenc}
\usepackage{amsmath, amssymb, graphicx}
\usepackage{algorithm}
\usepackage[noend]{algorithmic}
\usepackage{paralist}
\usepackage{dsfont}
\usepackage{graphicx}
\usepackage[caption = false]{subfig}
\usepackage{floatflt}
\usepackage{float}

\newcommand{\R}{\mathbb{R}}
\newcommand{\Normal}{\mathcal{N}}

\newcommand{\Order}{\mathcal{O}}

\newcommand{\bmat}{\begin{pmatrix}}
\newcommand{\emat}{\end{pmatrix}}

\begin{document}

\title{Challenges in High-dimensional Reinforcement Learning\\with Evolution Strategies}
\author{Nils M{\"u}ller and Tobias Glasmachers\\Institut f{\"u}r Neuroinformatik, Ruhr-Universit\"at Bochum, Germany\\\texttt{\{nils.mueller, tobias.glasmachers\}@ini.rub.de}}
\date{}

\maketitle

\begin{abstract}
Evolution Strategies (ESs) have recently become popular for training deep
neural networks, in particular on reinforcement learning tasks, a special
form of controller design.
Compared to classic problems in continuous direct search, deep networks
pose extremely high-dimensional optimization problems, with many thousands
or even millions of variables. In addition, many control problems give rise
to a stochastic fitness function. Considering the relevance of the
application, we study the suitability of evolution strategies for
high-dimensional, stochastic problems.
Our results give insights into which algorithmic mechanisms of modern ES 
are of value for the class of problems at hand, and they reveal principled
limitations of the approach. They are in line with our theoretical
understanding of ESs. We show that combining ESs that offer reduced
internal algorithm cost with uncertainty handling techniques yields
promising methods for this class of problems.
\end{abstract}

\section{Introduction}
\label{section:introduction}

Since the publication of DeepMind's Deep-Q-Learning system
\cite{mnih2015human} in 2015, the field of (deep) reinforcement learning
(RL) \cite{sutton1998reinforcement} is developing at a rapid pace.
In \cite{mnih2015human} neural networks learn to play classic Atari 2600
games solely from interaction, based on raw (unprocessed) visual input.
The approach had a considerable impact because it demonstrated the great
potential of deep reinforcement learning.
Only one year later AlphaGo \cite{silver2016mastering} demystified
the ancient game of Go by beating multiple human world experts.
In this rapidly moving field, Evolution Strategies (ESs)
\cite{beyer2002evolution,hansen2001completely,rechenberg1973evolutionsstrategie}
have gained considerable attention by the machine learning community
when OpenAI promoted them as a ``scalable alternative to reinforcement
learning''~\cite{salisman2017evolution}, which spawned
several follow-up works \cite{chrabaszcz2018back,such2017deep}.

Already long before deep RL, controller design with ESs was studied for
many years within the domain of neuroevolution
\cite{geijtenbeek2013flexible,heidrich2009neuroevolution,igel2003neuroevolution,moriarty1999evolutionary,stanley2009hypercube,stanley2002evolving}.
The optimization of neural network controllers is frequently cast as an
episodic RL problem, which can be solved with direct policy search, for
example with an ES. This amounts to parameterizing a class
of controllers, which are optimized to maximize reward or to minimize
cost, determined by running the controller on the task at hand, often in a
simulator. The value of the state-of-the-art covariance matrix adaptation
evolution strategy (CMA-ES) algorithm \cite{hansen2001completely} for this
problem class was emphasized by several authors \cite{geijtenbeek2013flexible,heidrich2009neuroevolution}.
CMA-ES was even augmented with an uncertainty handling mechanism,
specifically for controller design~\cite{hansen2009method}.

The controller design problems considered in the above-discussed papers
are rather low-dimensional, at least compared to deep learning models
with up to millions of weights. CMA-ES is rarely applied to problems
with more than 100 variables. This is because learning a full covariance
matrix introduces non-trivial algorithm internal cost and hence prevents
the direct application of CMA-ES to high-dimensional problems. In recent
years it turned out that even covariance matrix adaptation can be scaled
up to very large dimensions, as proven by a series of algorithms
\cite{akimoto2014comparison,loshchilov2017limited,loshchilov2014computationally,ros2008simple}, either by restricting the covariance matrix to the diagonal, to a
low-rank model, or to a combination of both. Although apparently
promising, none of these approaches was to date applied to the problem
of deep reinforcement learning.

Against this background, we investigate the suitability of evolution
strategies in general and of modern scalable CMA-ES variants in
particular for the design of large-scale neural network controllers.
In contrast to most existing studies in this domain, we approach the
problem from an optimization perspective, not from a (machine) learning
perspective. We are primarily interested in how different algorithmic
components affect optimization performance in high-dimensional, noisy
optimization problems. Our results provide a deeper understanding of
relevant aspects of algorithm design for deep neuroevolution.

The rest of the paper is organized as follows. After a brief
introduction to controller design we discuss mechanisms of evolution
strategies in terms of convergence properties.
We carry out experiments on RL problems as well as on optimization
benchmarks, and close with our conclusions.

\section{Problems under Study}

\paragraph{General setting.}
In this paper, we investigate the utility of evolution strategies for
optimization problems that pose several difficulties at the same time:
\begin{compactitem}
\item
	a large number $d$ of variables (high dimension of the search space
	$\R^d$),
\item
	fitness noise, i.e., the variability of fitness values $f(x)$ when
	evaluating the non-deterministic fitness function multiple times in
	the same point $x$, and
\item
	multi-modality, i.e., the presence of a large number of local optima.
\end{compactitem}
Additionally, a fundamental requirement of relatively quick problem
evaluation time (typically requiring simulation of real world phenomena)
is appropriate.

State-of-the-art algorithms like CMA-ES can handle dimensions
of up to $d \leq 100$ with ease. They become painfully slow for
$d \geq 1000$ due to their time and memory requirements.
In this sense, a high-dimensional problem is characterized by
$d \geq 1000$. Yet, recent advances led to the successful
application of ESs with a diagonal and/or low-rank model of
the covariance matrix to up to 500,000-dimensional (noise-free) problems
\cite{loshchilov2014computationally}. Mainly fueled by the reduction of
internal algorithm cost, modern ESs thereby become applicable to new
classes of problems. Deep reinforcement learning problems present such a
new challenge, characterized by a combination of three aspects, namely
high search space dimension, fitness noise, and multi-modality. While
neural networks are known to give rise to highly multi-modal landscapes,
several recent studies suggest that many if not all local optima are of
good or even equal quality \cite{kawaguchi2016deep}. Furthermore, the
problem can be addressed effectively with simple generic methods like
restarts.
Therefore we focus on the less well understood interaction of noise and
high dimensions.
As several components of modern ESs are impaired by uncertainty and sparsity
in sampling, their merit---especially as with increasing dimension the
relative share of function evaluations becomes prevalent in time---needs to
be assessed. To this end, we draw from previous work on uncertainty handling
\cite{beyer2017analysis,hansen2009method} in order to face fundamental
challenges like a low signal-to-noise ratio.

Despite the greater generality of the described problem setting, a
central motivation for studying the above problem class is controller
design. In evolutionary controller design, an individual (a candidate
controller) is evaluated in a Monte Carlo manner, by sampling its
performance on a (simulated) task, or a set of tasks and conditions.
Stochasticity caused by random state transitions and randomized
controllers is a common issue. Due to complex and stochastic
controller-environment interactions, controller design is considered a
difficult problem, and black-box approaches like ESs are well suited
for the task, in particular, if gradients are not available.

\paragraph{Reinforcement learning.}
In reinforcement learning, control problems are typically framed as
stochastic, time-discrete, Markov decision processes
$(S, A, P_{\cdot, \cdot}(\cdot), \allowbreak R_{\cdot}(\cdot, \cdot), \gamma)$
with the notion of a (software) agent embedded in an environment.
The agent ought to take an action $a \in A$ when presented with a
state $s\in S$ of the environment in order to receive a reward
$(s, s',a) \mapsto R_s(s',a) \in \mathbb{R}$ for a resulting state
transition to new state $s' \in S$ in the next time step. An individual
encodes a (possibly randomized) controller or policy $\pi_\theta: S \to A$
with parameters $\theta \in \Theta$, which is followed by the agent. It is
assumed that each policy yields a constant expected cumulative reward over
a fixed number of actions $\tau$ taken when acting according to it, as
the state transition probability
$(s, s', a) \mapsto P_{s, a}(s') = Pr(\mathbf{s'}=s'|\mathbf{s}=s, \mathbf{a}=a)$,
to a successor state $s'$ is stationary (time-independent) and depends only on
the current state and action (Markov property), for all $s, s' \in S, a \in A$.
This cumulative reward acts as a fitness measure $F_\pi:\Theta \to \mathbb{R}$,
while the policy parameters $\theta$ (e.g., weights of neural networks
$\pi_\theta$) are the variables of the problem.
Thus, we consider the (reinforcement learning) optimization problem
\begin{equation*}
\min\limits_{\theta \in \Theta} \,\, F_{\pi}(\theta) =
-\!\!\!\!\!\!\!\displaystyle\sum_{s_{0},\dots,s_{\tau}\in S}\left(\sum_{k=0}^{\tau-1} \gamma^{k}R_{s_{k}}(s_{k+1},\pi_\theta(s_{k}))\right)\cdot \left(\prod_{j=0}^{\tau-1} P_{s_{j}, \pi_\theta(s_{j})}(s_{j+1})\right),
\label{eq:objective}
\end{equation*}
where $\gamma \in (0, 1]$ is a discount factor.

Developments in RL demonstrated the merit in utilizing ``model-free''
approaches to the design of high-dimensional controllers such as neural
networks for solving a variety of tasks previously inaccessible
\cite{silver2016mastering,mnih2015human}, as well as novel frameworks
for scaling evolution strategies to CPU clusters
\cite{salisman2017evolution}.

ESs have advantages and disadvantages compared to alternative approaches
like policy gradient methods. Several mechanisms of ESs add robustness
to the search. Modeling distributions over policy parameters as done
explicitly in natural evolution strategies (NES) \cite{wierstra2014natural}
and also in CMA-ES serves this purpose \cite{lehman2017more}, and so
do problem-agnostic algorithm design and strong invariance properties.
Direct policy search does not suffer from the temporal credit assignment
problem or from sparse rewards \cite{salisman2017evolution}. ESs have
demonstrated superior exploration behavior, which is important to avoid
a high bias when sampling the environment
\cite{plappert2017exploration}. On the contrary, ESs ignore the
information contained in individual state transitions and rewards. This
inefficiency can (partly) be compensated by better parallelism in
ESs~\cite{salisman2017evolution}.

\section{Evolution Strategies}
\label{section:ES}

In this section, we discuss Evolution Strategies (ESs) from a bird's eye
perspective, in terms of their central algorithmic components, and without
resorting to the details of their implementation. For actual exemplary
algorithm instances with the properties under consideration, we refer to
the literature. Algorithm~\ref{algo:ES} summarizes commonly found mechanisms
without going into any details.

\begin{algorithm*}[tb!]
\caption{Generic Evolution Strategy Template}
\label{algo:ES}
\begin{algorithmic}[1]
\STATE {initialize $\lambda$, $m \in \R^d$, $\sigma > 0$, $C = I$}
\REPEAT
  \REPEAT
    \FOR{$i \leftarrow 1,\ldots,\lambda$}
      \STATE {sample offspring $x_i \sim \Normal(m, \sigma^2 C)$}
      \STATE {evaluate fitness $f(x_i)$ by testing the controller encoded by $x_i$ on the task}
    \ENDFOR
    \STATE {actual optimization: update $m$}
    \STATE {step size control: update $\sigma$}
    \STATE {covariance matrix adaptation: update $C$}
    \STATE {uncertainty handling, i.e., adapt $\lambda$ or the number of tests per fitness evaluation}
  \UNTIL{ \textit{stopping criterion is met} }
  \STATE {prepare restart, i.e., set new initial $m$, $\sigma$, and $\lambda$, and reset $C \leftarrow I$}
\UNTIL{ \textit{budget is used up} }
\STATE {return $m$}
\end{algorithmic}
\end{algorithm*}

ESs enjoy many invariance properties. This is generally considered a
sign of good algorithm design: due to their rank-based processing of
fitness values, they are invariant to strictly monotonically
increasing transformations of fitness; furthermore, they are invariant
to translation, rotation, and scaling provided that the initial
distribution is transformed accordingly, and with CMA (see below) they
are even asymptotically invariant under arbitrary affine
transformations.

\paragraph{Step Size Control.}
The algorithms applied to RL problems in
\cite{chrabaszcz2018back,such2017deep,salisman2017evolution} are
designed in the style of non-adaptive algorithms, i.e., applying a
mutation distribution with fixed parameters $\sigma$ and $C$, adapting
only the center $m$. This method is known to converge as slowly as
pure random search~\cite{hansen2015evolution}. Therefore it is in
general imperative to add step size adaptation, which has always
been an integral mechanism since the inception of the method
\cite{beyer2002evolution,rechenberg1973evolutionsstrategie}.
Cumulative step size adaptation (CSA) is a state-of-the-art method
\cite{hansen2001completely}. Step size
control enables linear convergence on scale invariant (e.g., convex
quadratic) functions, and hence locally linear convergence into twice
continuously differentiable local optima \cite{hansen2015evolution},
which puts ESs into the same class as many gradient-based methods. It
was shown in \cite{teytaud2006general} that convergence of rank-based
algorithms cannot be faster than linear. However, the convergence rate
of a step size adaptive ESs is of the form $\Order(1/(kd))$, where $d$
is the dimensionality of the search space and $k$ is the condition
number of the Hessian in the optimum. In contrast, the convergence rate
of gradient descent suffers from large $k$, but is independent of the
dimension $d$.

\paragraph{Metric Learning.}
Metric adaptation methods like CMA-ES
\cite{beyer2017simplify,wierstra2014natural,hansen2001completely}
improve the convergence rate to $\Order(1/d)$ by adapting not only the
global step size $\sigma$ but also the full covariance matrix $C$ of
the mutation distribution. However, estimating a suitable covariance
matrix requires a large number of samples, so that fast progress is made
only after $\Order(d^2)$ fitness evaluations,
which is in itself prohibitive for large $d$. Also, the algorithm
internal cost for storing and updating a full covariance matrix and even
for drawing a sample from the distribution is at least of order
$\Order(d^2)$, which means that modeling the covariance matrix quickly
becomes the computational bottleneck, in particular if the fitness
function scales linear with $d$, as it is the case for neural networks.

Several ESs for large-scale optimization have been proposed as a remedy
\cite{akimoto2014comparison,loshchilov2017limited,sun2013linear,loshchilov2014computationally,ros2008simple}.
They model only the diagonal of the covariance matrix and/or interactions in an
$\Order(1)$ to $\Order(\log(d))$ dimensional subspace, achieving a time
and memory complexity of $\Order(d)$ to $\Order(d \log(d))$ per sample.
The aim is to offer a reasonable compromise between ES-internal and external
(fitness) complexity while retaining most of the benefits of full covariance
matrix adaptation. The LM-MA-ES algorithm \cite{loshchilov2017limited} offers
the special benefit of adapting the fastest evolving subspace of the covariance
matrix with only $\Order(d)$ samples, which is a significant speed-up over the
$\Order(d^2)$ sample cost of full covariance matrix learning.

\paragraph{Noise Handling.}
Evolution strategies can be severely impaired by noise, in particular
when it interferes with step size adaptation. Being randomized algorithms,
ESs are capable of tolerating some level of noise with ease.
In the easy-to-analyze multiplicative noise model \cite{jebalia2008multiplicative},
the noise level decays as we approach the optimum and hence, on the sphere
function $f(x) = \|x\|^2$, the signal-to-noise ratio (defined as the systematic variance of $f$ due to sampling divided by the variance of random noise)
oscillates around a positive constant (provided that step size
adaptation works as desired \cite{jagerskupper2006}, keeping $\sigma$
roughly proportional to $\|m\|/d$).
For strong noise, this ratio is small. Then the ES essentially performs
a random walk, and a non-elitist algorithm may even diverge. Then CSA is
endangered to converge prematurely \cite{beyer2003qualms}.
For more realistic additive noise, the noise variance is (lower bounded by)
a positive constant. When converging to the optimum, $\sigma$ and hence
the signal-no-noise-ratio decays to zero. Therefore progress stalls at some
distance to the optimum. Thus there exists a principled limitation on the
precision to which an optimum can be located.
Explicit noise handling mechanisms like \cite{beyer2017analysis,hansen2009method}
can be employed to increase the precision, and even enable convergence,
e.g., by adaptively increasing the population size or the number of
independent controller test runs per fitness evaluation. They 
adaptively increase the population size or the number of simulation
runs per fitness evaluation, effectively improving the signal-to-noise
ratio. The algorithm parameters can be tuned to avoid premature
convergence of CSA. However, the convergence speed is so slow that in
practice additive noise imposes a limit on the attainable solution
precision, even if the optimal convergence rate is
attained~\cite{beyer2017analysis}.

\paragraph{Noise in High Dimensions.}
There are multiple ways in which optimization with noise and
in high dimensions interact. In the best case, adaptation slows down
due to reduced information content per sample, which is the case for
metric learning. The situation is even worse for step size adaptation:
for the noise-free sphere problem,
the optimal step size $\sigma$ is known to be proportional to $\|m\|/d$.
Therefore, in the same distance to the optimum and for the same noise
strength, noise handling becomes harder in high dimensions.
Then the step size can become too small, and CSA can converge prematurely
\cite{beyer2003qualms}.

\section{Experiments}
\label{section:experiments}

Most of the theoretical arguments brought forward in the previous
section are of asymptotic nature, while sometimes practice is dominated
by constant factors and transient effects. Also, it remains unclear
which of the different effects like slow convergence, slow adaptation,
and the difficulty of handling noise is a critical factor. In this
section, we provide empirical answers to these questions.

Well-established benchmark collections exist in the evolutionary
computation domain, in particular for continuous search spaces
\cite{hansen2016coco,li2013benchmark}. Typical questions are whether an
algorithm can handle non-separable, ill-conditioned, multi-modal, or
noisy test functions. However, it is not a priori clear which of these
properties are found in typical controller design problems. For example,
the optimization landscapes of neural networks are not yet well
understood. Closing this gap is far beyond the scope of this paper. Here
we pursue a simpler goal, namely to identify the most relevant factors.
More concretely, we aim to understand in which situation (dimensionality
and noise strength) which algorithm component (as discussed in the
previous section) has a significant impact on optimization performance,
and which mechanisms fail to pay off.

To this end, we run different series of experiments on established
benchmarks from the optimization literature and from the RL domain.
We have published code for reproducing all experiments online.%
\footnote{\texttt{https://github.com/NiMlr/High-Dim-ES-RL}}
For ease of comparison, we use the recently proposed MA-ES algorithm
\cite{beyer2017simplify} adapting the full covariance matrix, which was
shown empirically to perform very similar to CMA-ES. This choice is
motivated by its closeness to the LM-MA-ES method
\cite{loshchilov2017limited}, which learns a low-rank approximation of
the covariance matrix. When disabling metric learning entirely in
these methods, we obtain a rather simple ES with CSA, which we include
in the comparison.

\begin{figure}
\caption{Evolution of population average fitness for three reinforcement learning
	tasks with LM-MA-ES, averaged over five runs.
	\label{figure:fitness-RL}
}
\begin{center}
\includegraphics[width=0.9\textwidth,height=3.75cm]{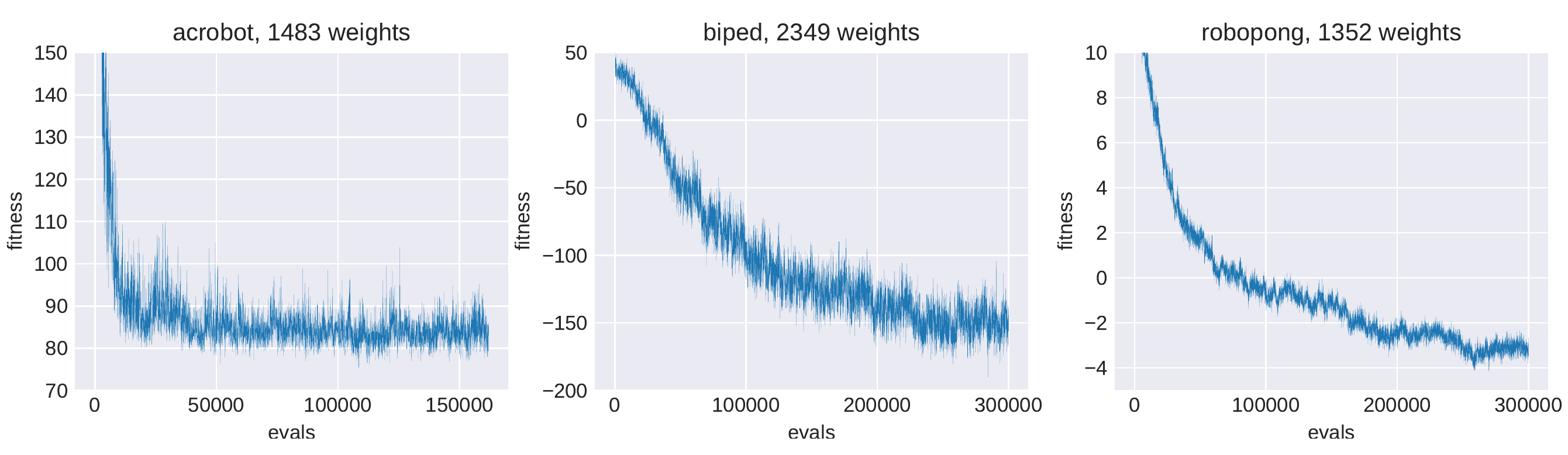}
\end{center}
\vspace*{-8mm}
\end{figure}

Figure~\ref{figure:fitness-RL} shows the time evolution of the fitness $F_\pi(\theta)$ (eq.~\eqref{eq:objective}) on
three prototypical benchmark problems from the OpenAI Gym environment
\cite{OpenAiGym}, a collection of RL benchmarks: acrobot, bipedal walker,
and robopong. All three controllers $\pi_\theta$ (eq.~\eqref{eq:objective}) are fully connected deep networks
with hidden layer sizes 30-30-10 (acrobot) and 30-30-15-10 (bipedal
walker and robopong), giving rise to moderate numbers of weights around 2,000,
depending on the task-specific numbers of inputs and controls.
It is apparent that in all three cases fitness noise plays a key role.

\begin{figure}
\caption{Evolution of fitness or neural networks with different
	numbers of weights (different hidden layer sizes), for LM-MA-ES
    (left) and MA-ES (right) on the bipedal walker task.
	\label{figure:fitness-scaling}
}
\begin{center}
\includegraphics[width=0.8\textwidth,height=5cm]{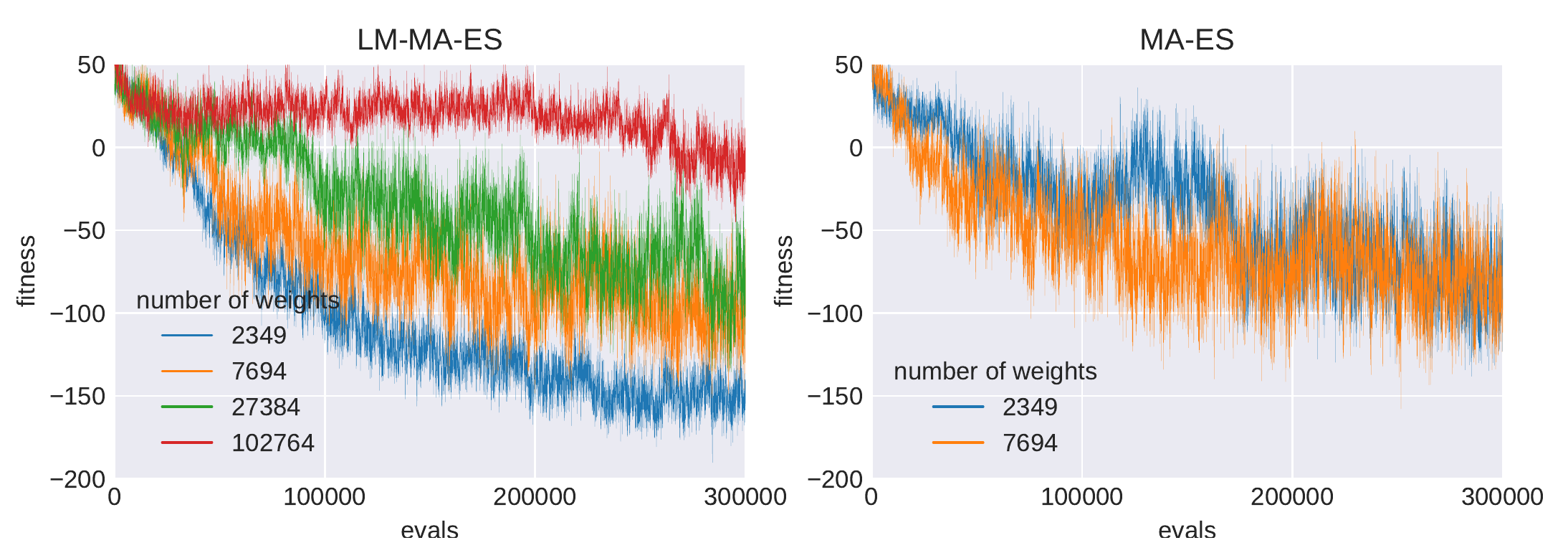}
\end{center}
\vspace*{-8mm}
\end{figure}

Figure~\ref{figure:fitness-scaling} investigates the scaling of LM-MA-ES
and MA-ES with problem dimension on the bipedal walker task. For the
small network considered above, MA-ES performs considerably worse than
LM-MA-ES, not only in wall clock time (not shown) but also in terms of
sample complexity. A similar effect was observed in
\cite{loshchilov2017limited} for the Rosenbrock problem. This indicates
that LM-MA-ES can profit from its fast adaptation to the most prominent
subspace. However, this effect does not necessarily generalize to other
tasks. More importantly, we see (unsurprisingly) that the performance of
both algorithms is severely affected as $d$ grows.

In order to gain a better understanding of the effect of fitness noise
on high-dimensional controller design, we consider optimization benchmarks.
These problems have the advantage that the optimum is known and that the
noise strength is under our control. Since we are particularly
interested in scalable metric learning, we employ the noisy ellipsoid problem
$f(x) = \bar f(x) + N(x)$, $\bar f(x) = \sqrt{x^T H x}$, with eigenvalues
$\lambda_i = k^{\frac{i-1}{d-1}}$ of $H$, and $N(x)$ is the noise. For the
multiplicative case, the range of $N(x)$ is proportional to $\bar f(x)$,
while for the additive case it is not.

\begin{figure}
\caption{Evolution of fitness and step size over function evaluations,
	averaged over five independent runs, for three different algorithms
	and problems (see the legend for details). Note the logarithmic
	scale of both axes.
	\label{figure:opt-results}
}
\begin{center}
\includegraphics[width=\textwidth,height=8.85cm]{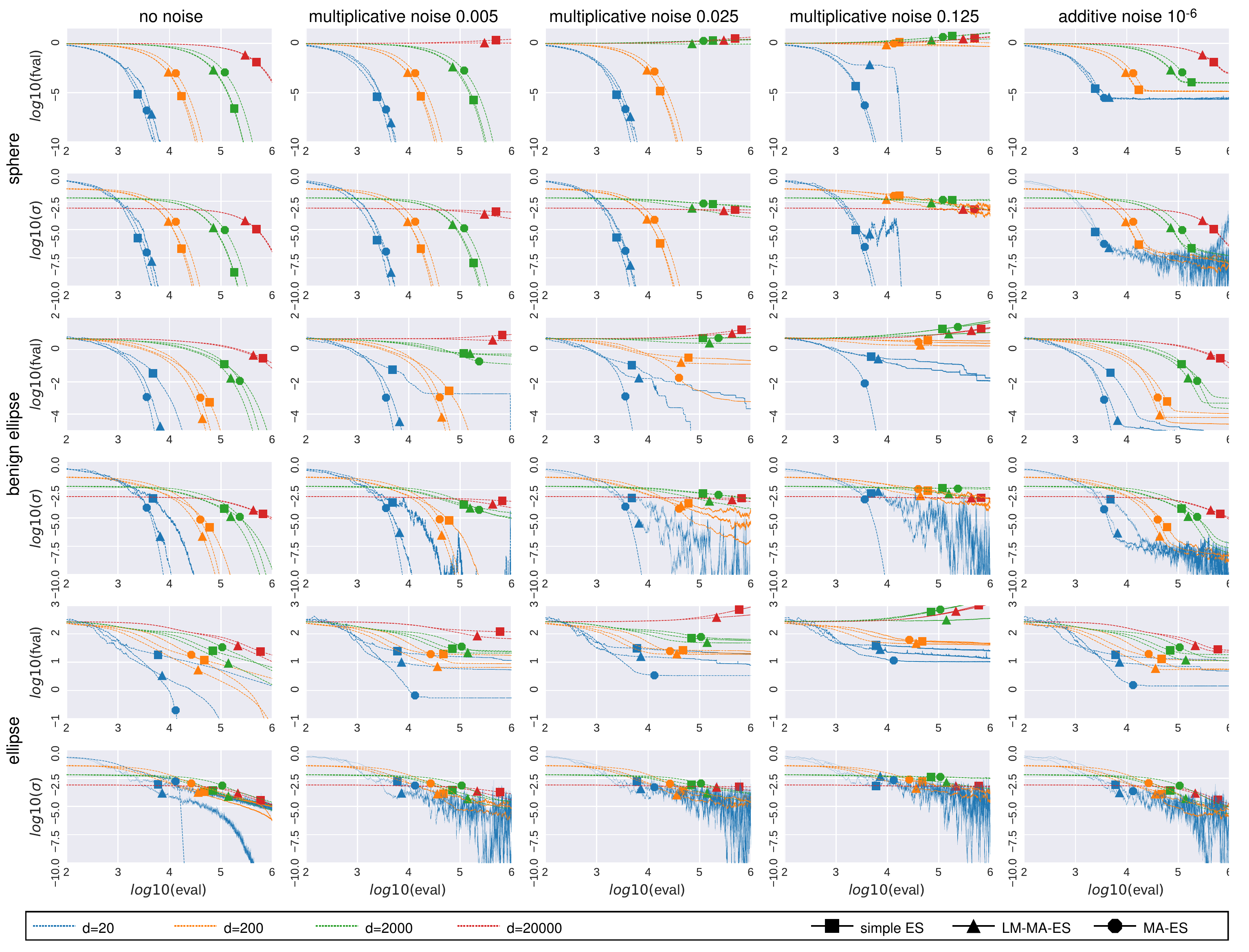}
\end{center}
\vspace*{-8mm}
\end{figure}

Among the problem parameters we vary
\begin{compactitem}
\item problem dimension $d \in \{20, 200, 2000, 20000\}$,
\item problem conditioning ($k \in \{10^0, 10^2, 10^6\}$ (sphere, benign ellipsoid, standard ellipsoid), and
\item noise strength (none, multiplicative with various constants of proportionality, additive).
\end{compactitem}
Figure~\ref{figure:opt-results} shows the time evolution of fitness
and step size of the different algorithms in these conditions.

The experiments on the noise-free sphere problem show that the speed of
optimization decays with increasing dimension, as predicted by theory
\cite{jagerskupper2006}: halving the distance to the optimum requires
$\Theta(d)$ samples. For this reason, within the fixed budget of $10^6$
function evaluations, there is less progress in higher dimensions. For
$d=20,000$, the solution quality is still improved by a factor of about
$10^3$, which requires the step size to change by the same amount.
However, extrapolating our results we see that in extremely high
dimensions the algorithm is simply not provided enough generations to make
sufficient progress in order to justify step size adaptation. This is in
accordance with \cite{chrabaszcz2018back}.
A similar effect is observed for metric learning, which takes
$\Theta(d^2)$ samples for the full covariance matrix. Even for the still
moderate dimension of $d=2,000$, the adaptation process is not completed
within the given budget. Yet, also during the transitional phase where
the matrix is not yet fully adapted, MA-ES already has an edge over the
simple ES. LM-MA-ES is sometimes better and sometimes worse than MA-ES.
It may profit from the significantly smaller number of parameters in the
low-rank covariance matrix, which allows for faster adaptation, in
particular in high dimensions, where MA-ES does not have enough samples to
complete its learning phase. In any case, its much lower internal
complexity allows us to scale up LM-MA-ES to much higher dimensions.

In summary, metric adaptation is still useful for problems with a ``realistic'' dimension of even very detailed controller design problems
in engineering, while it is too slow for training neural networks with
millions of weights, unless the budget grows at least linear with the
number of weights. This in turn requires extremely fast simulations as
well as a large amount of computational hardware resources.

Noise has a significant impact on the optimization behavior and on the
solution quality. Additive noise implies extremely slow convergence, and
indeed we find that all methods stall in this case. Too strong
multiplicative noise even results in divergence. A particularly
adversarial effect is that the noise strength that can be tolerated is at best
inversely proportional to the dimension. This effect nicely shows up in
the noisy sphere results. Here, uncertainty handling can help in principle,
since it improves the signal-to-noise adaptively to the needs of the
algorithm, but at the cost of more function evaluations per generation,
which amplifies the effects discussed above.

In the presence of noise, CSA does not seem to work well in low
dimensions. In case of high noise, $\log(\sigma)$ performs a random walk.
However, this walk is subject to a selection bias away from high values,
since they improve the signal-to-noise ratio. Therefore we find extended
periods of stalled progress, in particular for $d=20$, accompanied by a
random walk of the (far too small) step size. The effect is unseen in
higher dimensions, probably due to the smaller update rate.

\begin{floatingfigure}[r]{0.5\textwidth}
\caption{
	(UH-)LM-MA-ES on the benign ellipse in $d=100,000$ with
    additive noise restricted to $\bar f(x) > 3.5$. LM-MA-ES without
    uncertainty handling (blue curve) diverges while LM-MA-ES with
    uncertainty handling approaches the optimum (red curve).
	\label{figure:UH}
}
\begin{center}
\vspace*{-1em}
\includegraphics[width=0.45\textwidth]{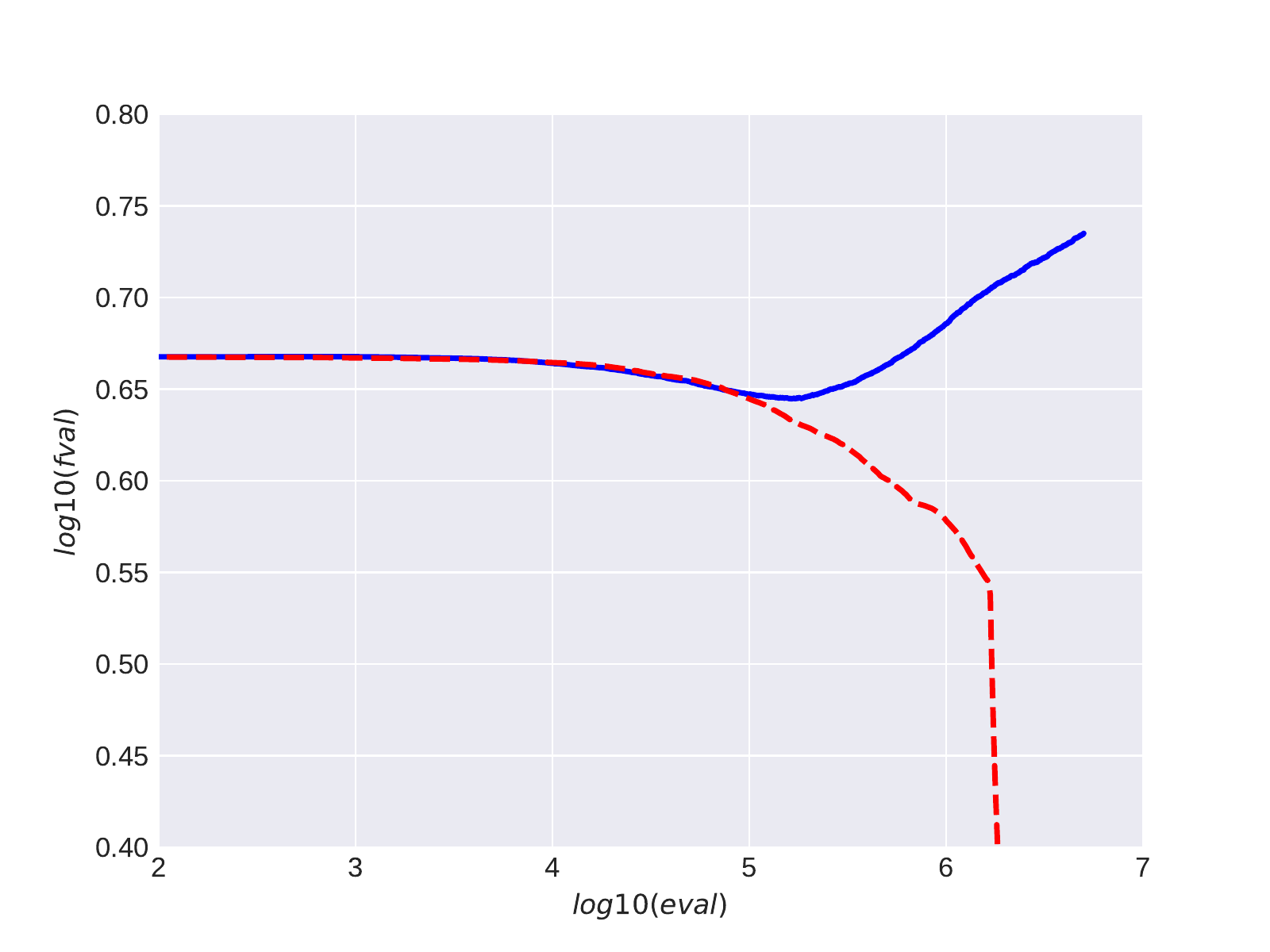}
\vspace*{-1em}
\end{center}
\end{floatingfigure}

We are particularly interested in the interplay between metric
adaptation and noise. It turns out that in all cases where CMA helps
(non-spherical problems of moderate dimension), i.e., where LM-MA-ES and
MA-ES outperform the simple ES, the same holds true for the
corresponding noisy problems. We conclude that metric learning still
works well, even when faced with noise in high dimensions.

The influence of noise can be controlled and mitigated with uncertainty
handling techniques \cite{beyer2017analysis,hansen2009method}.
This essentially results in curves similar to the
leftmost column of figure~\ref{figure:opt-results}, but with slower
convergence, depending on the noise strength. In controller design,
noise handling can be key to success, in particular if the optimal
controller is nearly deterministic, while strong noise is encountered
during learning. This is a plausible assumption for the bipedal walker
task: at an intermediate stage, the walker falls over randomly depending
on minor details of the environment, resulting in high noise variance,
while a controller that has learned a stable and robust walking pattern
achieves good performance with low variance. Then it is key to handle
the early phase by means of uncertainty handling, which enables the ES
to enter the late convergence phase eventually. Figure~\ref{figure:UH}
displays such a situation for the benign ellipse with $d=100,000$ with
additive noise applied only for function values above a threshold.
LM-MA-ES without uncertainty handling fails, but with uncertainty
handling the algorithm finally reaches the noise-free region and then
converges quickly.

\begin{figure}[t]
\caption{Fitness and number of re-evaluations (left) step size and standard deviation of fitness (right), averaged over six runs of LM-MA-ES with and without uncertainty handling on the bipedal walker task.
	\label{figure:RL-UH}
}
\begin{center}
\includegraphics[width=\textwidth]{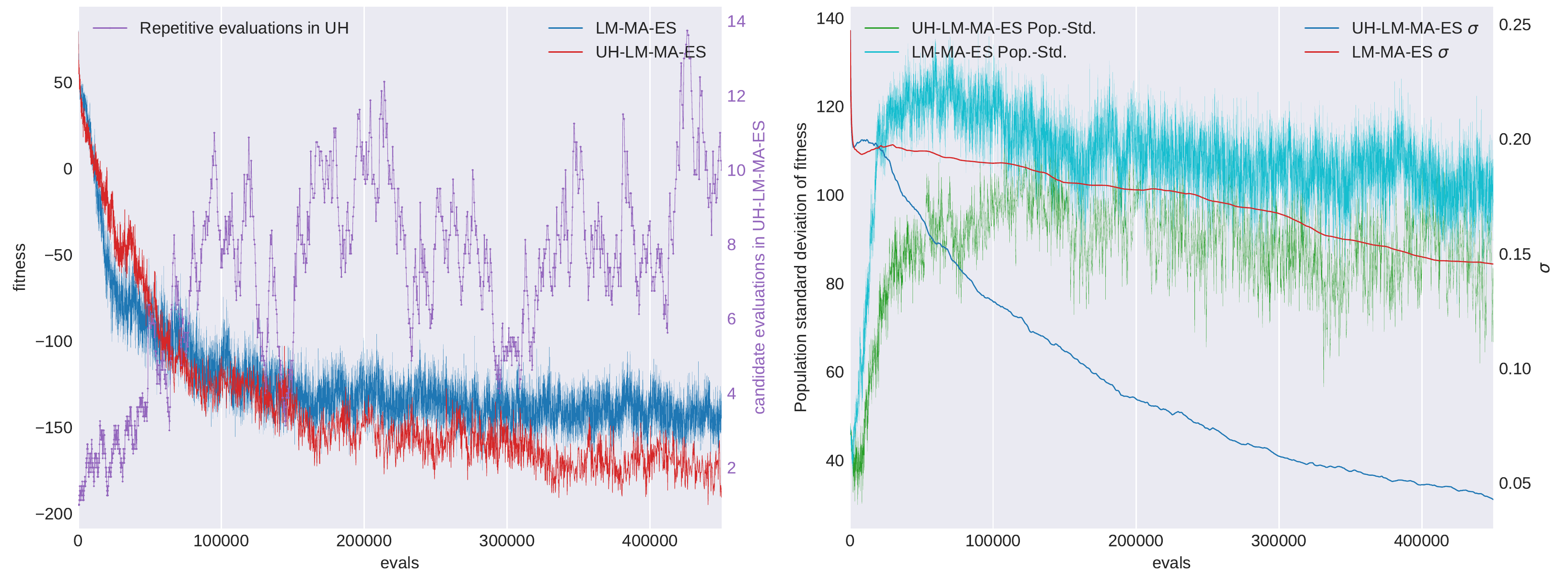}
\vspace*{-4
em}
\end{center}
\end{figure}

Figure~\ref{figure:RL-UH} shows the effect of uncertainty handling.
It yields significantly more stable optimization behavior in two ways: 1. it keeps the
step size high, avoiding an undesirable decay and hence the danger of
premature convergence or of a less-robust population, and 2. it keeps the fitness variance small,
which allows the algorithm to reach better fitness in the late fine
tuning phase.
Interestingly, the ES without uncertainty handling is initially faster.
This can be mitigated by tuning the initial step size, which anyway
becomes an increasingly important task in high dimensions, for two
reasons: adaptation takes long in high dimensions, and even worse,
a too small initial step size makes uncertainty handling kick in without
need, so that the adaptation takes even longer. The latter might especially be called for on expensive problems commonly found in RL.

\section{Conclusion}
\label{section:conclusion}

We have investigated the utility of different algorithmic mechanisms of
evolution strategies for problems with a specific combination of
challenges, namely high-dimensional search spaces and fitness noise. The
study is motivated by a broad class of problems, namely the design of
flexible controllers. Reinforcement learning with neural networks yields
some extremely high-dimensional problem instances of this type.

We have argued theoretically and also found empirically that many of the
well-established components of state-of-the-art methods like CMA-ES and
scalable variants thereof gradually lose their value in high dimensions,
unless the number of function evaluations can be scaled up accordingly.
This affects the adaptation of the covariance matrix, and in extremely
high-dimensional cases also the step size. This somewhat justifies the
application of very simple algorithms for training neural networks with
millions of weights, see~\cite{chrabaszcz2018back}.

Additive noise imposes a principled limitation on the solution quality.
However, it turns out that adaptation of the search distribution still
helps, because it allows for a larger step size and hence a better
signal-to-noise ratio. Unsurprisingly, uncertainty handling can be a
key technique for robust convergence.

Overall, we find that adaptation of the mutation distribution becomes
less valuable in high dimensions because it kicks in only rather late.
However, it never harms, and it can help even when dealing with noise in
high dimensions. Our results indicate that a scalable modern evolution
strategy with step size and efficient metric learning equipped with
uncertainty handling is the most promising general-purpose technique for
high-dimensional controller design.

\bibliographystyle{plain}

{\small
\begin{thebibliography}{10}

\bibitem{akimoto2014comparison}
Youhei Akimoto, Anne Auger, and Nikolaus Hansen.
\newblock Comparison-based natural gradient optimization in high dimension.
\newblock In {\em Proceedings of the 2014 Annual Conference on Genetic and
  Evolutionary Computation}, pages 373--380. ACM, 2014.

\bibitem{beyer2003qualms}
Hans-Georg Beyer and Dirk~V Arnold.
\newblock Qualms regarding the optimality of cumulative path length control in
  {CSA}/{CMA}-evolution strategies.
\newblock {\em Evolutionary Computation}, 11(1):19--28, 2003.

\bibitem{beyer2017analysis}
Hans-Georg Beyer and Michael Hellwig.
\newblock Analysis of the {pcCMSA-ES} on the noisy ellipsoid model.
\newblock In {\em Proceedings of the Genetic and Evolutionary Computation
  Conference}, pages 689--696. ACM, 2017.

\bibitem{beyer2002evolution}
Hans-Georg Beyer and Hans-Paul Schwefel.
\newblock Evolution strategies--a comprehensive introduction.
\newblock {\em Natural computing}, 1(1):3--52, 2002.

\bibitem{beyer2017simplify}
Hans-Georg Beyer and Bernhard Sendhoff.
\newblock Simplify your covariance matrix adaptation evolution strategy.
\newblock {\em IEEE Transactions on Evolutionary Computation}, 2017.

\bibitem{chrabaszcz2018back}
Patryk Chrabaszcz, Ilya Loshchilov, and Frank Hutter.
\newblock Back to basics: Benchmarking canonical evolution strategies for
  playing atari.
\newblock Technical Report 1802.08842, arXiv.org, 2018.

\bibitem{wierstra2014natural}
Daan~Wierstra et~al.
\newblock {Natural Evolution Strategies}.
\newblock {\em Journal of Machine Learning Research}, 15(1):949--980, 2014.

\bibitem{silver2016mastering}
David~Silver et~al.
\newblock Mastering the game of {Go} with deep neural networks and tree search.
\newblock {\em Nature}, 529(7587):484--489, 2016.

\bibitem{such2017deep}
Felipe~Such et~al.
\newblock Deep neuroevolution: Genetic algorithms are a competitive alternative
  for training deep neural networks for reinforcement learning.
\newblock Technical Report 1712.06567, arXiv.org, 2017.

\bibitem{OpenAiGym}
Greg~Brockman et~al.
\newblock {OpenAI Gym}.
\newblock Technical Report 1606.01540, arxiv.org, 2016.

\bibitem{loshchilov2017limited}
Ilya~Loshchilov et~al.
\newblock Limited-memory matrix adaptation for large scale black-box
  optimization.
\newblock Technical Report 1705.06693, arXiv.org, 2017.

\bibitem{lehman2017more}
Joel~Lehman et~al.
\newblock {ES} is more than just a traditional finite-difference approximator.
\newblock Technical Report 1712.06568v2, arXiv.org, 2017.

\bibitem{plappert2017exploration}
Matthias~Plappert et~al.
\newblock Parameter space noise for exploration.
\newblock Technical Report 1706.01905v2, arXiv.org, 2017.

\bibitem{hansen2009method}
Nikolaus~Hansen et~al.
\newblock A method for handling uncertainty in evolutionary optimization with
  an application to feedback control of combustion.
\newblock {\em IEEE Transactions on Evolutionary Computation}, 13(1):180--197,
  2009.

\bibitem{hansen2016coco}
Nikolaus~Hansen et~al.
\newblock {COCO}: A platform for comparing continuous optimizers in a black-box
  setting.
\newblock Technical Report 1603.08785, arXiv.org, 2016.

\bibitem{geijtenbeek2013flexible}
Thomas~Geijtenbeek et~al.
\newblock Flexible muscle-based locomotion for bipedal creatures.
\newblock {\em ACM Transactions on Graphics (TOG)}, 32(6):206, 2013.

\bibitem{salisman2017evolution}
Tim~Salimans et~al.
\newblock Evolution strategies as a scalable alternative to reinforcement
  learning.
\newblock Technical Report 1703.03864, arXiv.org, 2017.

\bibitem{mnih2015human}
Volodymyr~Mnih et~al.
\newblock Human-level control through deep reinforcement learning.
\newblock {\em Nature}, 518(7540):529, 2015.

\bibitem{li2013benchmark}
Xiaodong~Li et~al.
\newblock Benchmark functions for the {CEC} 2013 special session and
  competition on large-scale global optimization.
\newblock {\em gene}, 7(33):8, 2013.

\bibitem{sun2013linear}
Yi~Sun et~al.
\newblock A linear time natural evolution strategy for non-separable functions.
\newblock In {\em Conference companion on genetic and evolutionary
  computation}. ACM, 2013.

\bibitem{hansen2015evolution}
Nikolaus Hansen, Dirk~V Arnold, and Anne Auger.
\newblock Evolution strategies.
\newblock In {\em Springer handbook of computational intelligence}, pages
  871--898. Springer, 2015.

\bibitem{hansen2001completely}
Nikolaus Hansen and Andreas Ostermeier.
\newblock Completely derandomized self-adaptation in evolution atrategies.
\newblock {\em Evolutionary Computation}, 9(2):159--195, 2001.

\bibitem{heidrich2009neuroevolution}
Verena Heidrich-Meisner and Christian Igel.
\newblock Neuroevolution strategies for episodic reinforcement learning.
\newblock {\em Journal of Algorithms}, 64(4):152--168, 2009.

\bibitem{igel2003neuroevolution}
Christian Igel.
\newblock Neuroevolution for reinforcement learning using evolution strategies.
\newblock In {\em Congress on Evolutionary Computation}, volume~4, pages
  2588--2595, 2003.

\bibitem{jagerskupper2006}
Jens J{\"a}gersk{\"u}pper.
\newblock How the {(1+1)-ES} using isotropic mutations minimizes positive
  definite quadratic forms.
\newblock {\em Theoretical Computer Science}, 361(1):38--56, 2006.

\bibitem{jebalia2008multiplicative}
Mohamed Jebalia and Anne Auger.
\newblock On multiplicative noise models for stochastic search.
\newblock In {\em Parallel Problem Solving from Nature}, pages 52--61.
  Springer, 2008.

\bibitem{kawaguchi2016deep}
Kenji Kawaguchi.
\newblock Deep learning without poor local minima.
\newblock In {\em Advances in Neural Information Processing Systems}, pages
  586--594, 2016.

\bibitem{loshchilov2014computationally}
Ilya Loshchilov.
\newblock A computationally efficient limited memory {CMA-ES} for large scale
  optimization.
\newblock In {\em Proceedings of the 2014 Annual Conference on Genetic and
  Evolutionary Computation}, pages 397--404. ACM, 2014.

\bibitem{moriarty1999evolutionary}
David~E Moriarty, Alan~C Schultz, and John~J Grefenstette.
\newblock Evolutionary algorithms for reinforcement learning.
\newblock {\em J. Artif. Intell. Res.(JAIR)}, 11:241--276, 1999.

\bibitem{rechenberg1973evolutionsstrategie}
Ingo Rechenberg.
\newblock {Evolutionsstrategie--Optimierung technischer Systeme nach Prinzipien
  der biologischen Evolution}.
\newblock 1973.

\bibitem{ros2008simple}
Raymond Ros and Nikolaus Hansen.
\newblock A simple modification in {CMA-ES} achieving linear time and space
  complexity.
\newblock In {\em International Conference on Parallel Problem Solving from
  Nature}, pages 296--305. Springer, 2008.

\bibitem{stanley2009hypercube}
Kenneth Stanley, David D'Ambrosio, and Jason Gauci.
\newblock A hypercube-based encoding for evolving large-scale neural networks.
\newblock {\em Artificial life}, 15(2):185--212, 2009.

\bibitem{stanley2002evolving}
Kenneth~O Stanley and Risto Miikkulainen.
\newblock Evolving neural networks through augmenting topologies.
\newblock {\em Evolutionary computation}, 10(2):99--127, 2002.

\bibitem{sutton1998reinforcement}
Richard~S Sutton and Andrew~G Barto.
\newblock {\em Reinforcement learning: An introduction}, volume~1.
\newblock MIT press Cambridge, 1998.

\bibitem{teytaud2006general}
Olivier Teytaud and Sylvain Gelly.
\newblock General lower bounds for evolutionary algorithms.
\newblock In {\em Parallel Problem Solving from Nature--PPSN IX}, pages 21--31.
  2006.

\end{thebibliography}
}

\end{document}